\documentclass{article}

    \PassOptionsToPackage{numbers, compress}{natbib}
\usepackage[final]{neurips_2021}
\usepackage{graphicx}
\usepackage[utf8]{inputenc} % allow utf-8 input
\usepackage[T1]{fontenc}    % use 8-bit T1 fonts
\usepackage{url}            % simple URL typesetting
\usepackage{booktabs}       % professional-quality tables
\usepackage{amsfonts}       % blackboard math symbols
\usepackage{nicefrac}       % compact symbols for 1/2, etc.
\usepackage{microtype}      % microtypography
\usepackage{xcolor}         % colors

\title{Automated Supervised Feature Selection for Differentiated Patterns of Care}

\author{%
  Catherine Wanjiru \\
  Carnegie Mellon University Africa\\
  \texttt{ccatheri@andrew.cmu.edu} \\
   \And
   William Ogallo\\
   IBM Research Africa \\
   \texttt{william.ogallo@ibm.com } \\
   \And
   Girmaw Abebe Tadesse \\
   IBM Research Africa \\
   \texttt{girmaw.abebe.tadesse@ibm.com } \\
   \And
   Charles Wachira \\
   IBM Research Africa \\
   \texttt{charles.wachira1@ibm.com} \\
   \And
   Isaiah Onando Mulang' \\
   IBM Research Africa\\
   \texttt{mulang.onando@ibm.com} \\
    \And
   Aisha Walcott-Bryant \\
   IBM Research Africa \\
   \texttt{alwalcott@ke.ibm.com} \\
}

\begin{document}

\maketitle

\begin{abstract}
  An automated feature selection pipeline was developed using several state-of-the-art feature selection techniques to select optimal features for Differentiating Patterns of Care (DPOC). The pipeline included three types of feature selection techniques; Filters, Wrappers and Embedded methods to select the top K features. Five different datasets with binary dependent variables were used and their different top K optimal features selected. The selected features were tested in the existing multi-dimensional subset scanning (MDSS) where the most anomalous subpopulations, most anomalous subsets, propensity scores, and effect of measures were recorded to test their performance. This performance was compared with four similar metrics gained after using all covariates in the dataset in the MDSS pipeline. We found out that despite the different feature selection techniques used, the data distribution is key to note when determining the technique to use.  
  
%   The abstract paragraph should be indented \nicefrac{1}{2}~inch (3~picas) on
%   both the left- and right-hand margins. Use 10~point type, with a vertical
%   spacing (leading) of 11~points.  The word \textbf{Abstract} must be centered,
%   bold, and in point size 12. Two line spaces precede the abstract. The abstract
%   must be limited to one paragraph.
\end{abstract}

\section{Introduction}
% \label{headings}
% - Variability of care
% - DPoC overview
% - Difficulty of studying variability across arbitrary numbers of covariates
% - Feature selection
% - Objective

Healthcare is characterized by large differences in disease patterns, patient response to interventions, and cost of care across patient subpopulations \cite{appleby2011variations,krumholz2013variations,senn2016mastering}. Some of the key challenges to understanding such non-random variations of care delivery and costs are complicated by the lack of appropriate approaches for analyzing complex interactions of factors and interventions captured in large-scale real-world evidence data. Conventional stratification and subgroup analyses approaches are highly manual, require domain expertise, and, more importantly, limits stakeholders to analyzing only a few variables beyond which it becomes computationally infeasible. Furthermore, these approaches lack a ‘data-driven knowledge discovery’ aspect as investigators must suggest beforehand which variables they would like to use in their analyses. Additionally, even though supervised machine learning approaches can be used to investigate variability in care, these approaches are either subject to several modeling assumptions and limitations or lack adequate interpretability \cite{mcfowland2018efficient}.

Fortunately, state-of-the-art subset scanning techniques from the anomalous pattern detection literature \cite{mcfowland2018efficient,neill2012fast,zhang2016identifying,somanchi2017} can be leveraged to enable principled, scalable, and unsupervised discovery of specific segments of a patient subpopulation that are anomalous. For example, our research team has developed a suite of functionalities for discovering and analyzing variations of care through automatic stratification and subgroup analysis across any arbitrary combination of features and interventions captured in observational healthcare data \cite{ogallo2021detection}. These subset scanning methods focus on identifying anomalous subsets of records in a multidimensional array that differ from expected behavior. The functionalities can be used to discover and analyze non-obvious segments of patient populations most significantly impacted by disease outcomes, costs of care, and related interventions across different application areas. They address the limitations of the state-of-the-art stratification and subgroup analysis approaches to enable fast and efficient search of the most anomalous subsets across exponentially many possible subsets of records in a dataset. Anomalousness is quantified using a scoring function, such as a log-likelihood ratio statistic, that is maximized over all subsets in a dataset to identify the subset with the highest score \cite{neill2012fast}. The scoring function exploits a mathematical property called Linear-Time Subset Scanning (LTSS) that allows the search to be conducted in linear time rather than exponential time \cite{neill2012fast}. Furthermore, to adjust for multiple hypothesis testing and estimate the statistical significance of identified anomalous subsets, randomization testing where parametric bootstrapping is used to estimate the empirical p-value of the subset \cite{mcfowland2013fast}.
 
Currently, however, subset scanning techniques primarily rely on manual domain-expert-driven feature selection before anomalous pattern detection. This limits the utility of the techniques in large-scale datasets.  Fortunately, several automated feature selection methods \cite{miao_2016_a} already exist in literature and can be leveraged to potentially improve downstream efficiency and interpretability of anomalous pattern detection tasks. However, there is a dearth of literature on the application of these feature selection techniques in anomalous pattern detection and the potential biases they introduced remain unknown.
 
The overarching goal of this study was to evaluate scalable supervised techniques for the automated selection of features for subsequent use in anomalous pattern detection. The specific objectives were twofold. First, we aimed to automatically identify the top-K candidate features to use for automated stratification and subgroup analysis given observational health data with an arbitrary number of features. Second, we aimed to compare the characteristics of the most anomalous subsets identified by subset scanning after applying different automated feature selection approaches. To this end, we developed a pipeline combining state-of-the-art supervised feature selection techniques with anomalous pattern detection and applied it to exemplar datasets. This paper reports our findings.

\section{Related Work} 
This study combines feature selection with subset scanning techniques from the anomalous pattern detection literature. Feature Selection is a dimensionality reduction technique that can be used to removing irrelevant, duplicate, or label leaking features from a large subset of features \cite{kumar_2015_a}. Feature selection can either be supervised or unsupervised. Supervised feature selection techniques include Filters, Wrappers, and Embedded techniques \cite{miao_2016_a}, while unsupervised techniques include techniques such as Principal Component Analysis \cite{kumar_2015_a}. In this study, Filters, Wrappers, and Embedded techniques are applied. Filters are used to remove non-useful features before modeling. This can be viewed as a preprocessing step where the relationships among features, as well as between features and the outcome variable are evaluated to pre-select features independent of any modeling algorithms \cite{molina_2002_feature}. In this study, we use measures such as the the Pearson correlation coefficient\cite{benesty2009pearson}, the Variance Inflation Factor \cite{o2007caution}, the Chi-Square test~\cite{mchugh2013chi}, and Normalized Mutual Information \cite{estevez2009normalized} for pre-selecting features. 

Wrapper methods use specific machine learning models to evaluate and select features \cite{miao_2016_a}. They measure the usefulness of features by learning a stepwise linear classifier or regressor using a specific variable selection method such as forward selection or backward elimination, and proceeding recursively until a stopping rule is reached. In this study, we use ordinary least squares regression with backward elimination to select the optimal $K$ features. Finally, the Embedded methods are feature selection techniques where the by-product of the model fitting is a feature importance ranking \cite{molina_2002_feature,miao_2016_a}. In embedded techniques, top K features can be selected from the ranking list returned by the model.

There are several subset scanning techniques in the anomalous pattern detection literature including Bias-Scan\cite{zhang2016identifying}, treatment effect subset scanning (TESS)\cite{mcfowland2018efficient}, and anomalous patterns of care (APC) Scan\cite{somanchi2017}. Our study specifically builds upon Bias-Scan\cite{zhang2016identifying}, an algorithm that was designed to on the discovery of the subpopulation with the most divergence between the true outcomes and the predicted probabilities of a binary classifier. Bias Scan analyzes tabular data with discrete/discretized covariates, a binary outcome, and predictions generated by a binary classifier. It maximizes a Bernoulli likelihood ratio scoring statistic over the exponentially many subsets of records in the data to identify the subset of records with the strongest evidence of the expected odds of the outcome being different from the predicted odds. To do so in linear time, Bias Scan exploits a priority function\cite{mcfowland2018efficient} that ranks the values of a given feature and then select the highest-scoring subset as the subset consisting of the ``top-k'' priority values for some $k \in [1, \ldots, J]$. In this study, we use a simplified version of Bias Scan that replaces the predictions of a classifier with a simple mean of the outcome across all records in the data.

\section{Methods} 

\begin{figure*}[ht]
  \centering
  \includegraphics[width=1\textwidth]{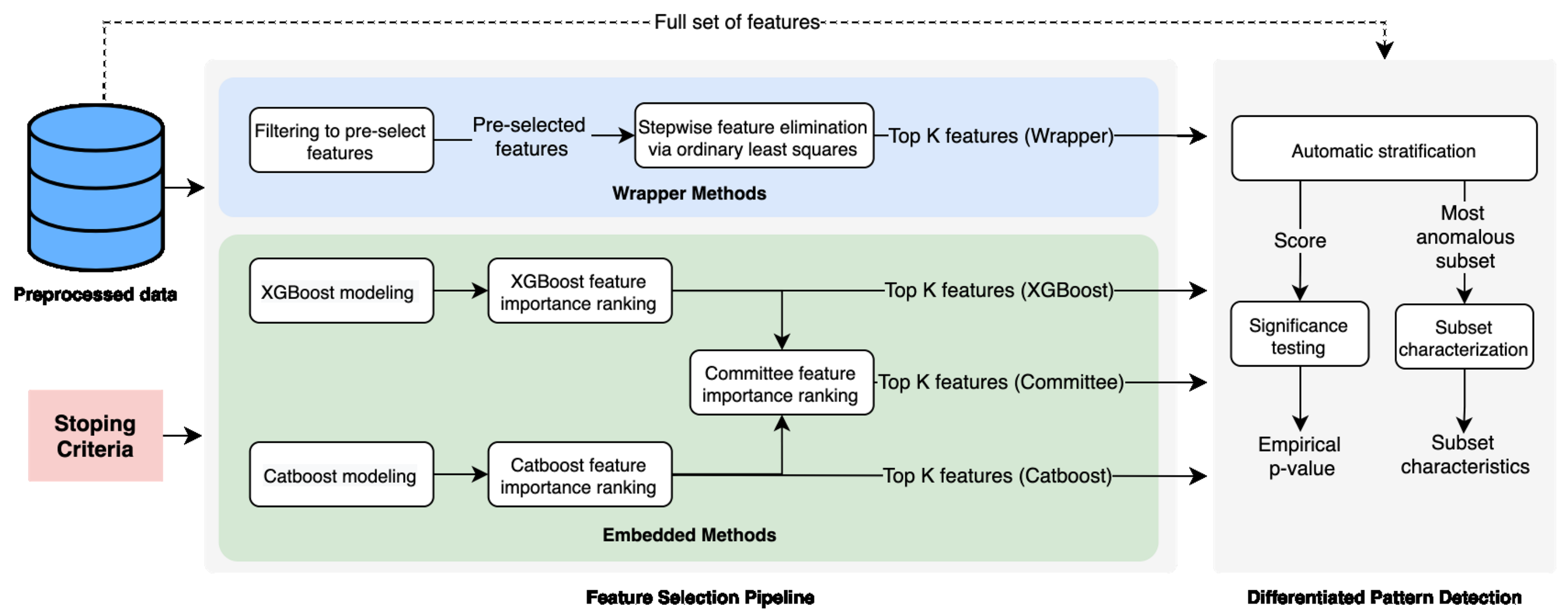}
  \caption{Pipeline combining feature selection and differentiated pattern detection.}
  \label{fig:pipeline}
\end{figure*}

Fig. \ref{fig:pipeline} illustrates our proposed pipeline for combining feature selection and differentiated pattern detection. The feature selection component takes as input processed data and user-specified stopping criteria (e.g. number of desired top K features) and uses different approaches to automatically select features from the supplied data. The differentiated pattern detection component scans over features selected by any of the different feature selection approaches to identify the highest-scoring (most anomalous) subsets. Next, we formulate the problem  mathematically and provide detailed descriptions of each component of the pipeline. 

% The workflow used in the project contained two major parts; the feature selection pipeline responsible for selecting the top k features and the existing MDSS pipeline to run experiments to test Effect Measures, Best Scores, Overlapping Subsets and Sub-populations for the selected k covariates.
% - Catherine (formalization - Girmaw)
% - diagram for the approach (William can help)
% - what was implemented (pipeline)
% - DPoC workflow/steps
% -Experiment:
%     - data (mimic
%     - covariates
%     - feature process
%     - DPoC
%     -Validations (with other datasets)

\subsection{Problem Formulation} 
Let $\mathcal{D}  = \{(x_i,y_i) | i = 1,2,\cdots, N\}$ denote a dataset containing $N$ samples where each sample $x_i$ is characterised by a set of $M$ features  $\mathcal{F}=[f_1,f_2,\cdots,f_M]$, and $y_i$ represents the outcome label.  The proposed automated feature selection process is defined as a function $\mathcal{R}(\cdot)$ that takes $\mathcal{D}$ and the required number of features $K$ as input and provides  $\mathcal{F}^r =\{f^r_1,f^r_2,\cdots,f^r_K\}$, where $K < M$, i.e., $\mathcal{D}^r=R(\mathcal{D},K) = \{(x^r_i,y_i) | i = 1,2,\cdots, N\}$ and $x^r_i$ is represented by $\mathcal{F}^r$.  Then  differentiated pattern detection process is defined as function $S(\cdot)$ that takes as input the selected features $\mathcal{F}^r$ and provides a subset of samples that are anomalous, $X^a=\{x^a_j\}_{j=1}^P$. Here, $P<N$, $X^a$ is represented by the identified anomalous feature values $\mathcal{F}^a=\{f^a_z\}_{z=1}^Z$, where $Z< K <M$ and $\hat{f^{a}_z}$ represents the value(s) of the $z^{th}$  feature in $\mathcal{F}^a$ that makes $X^a$ anomalous, e.g., $f^a_1= Gender$ and $\hat{f^a_1} = Female$. Note that $\mathcal{F}^a \subseteq \mathcal{F}^r \subseteq  \mathcal{F}$. The divergence of the identified subgroups $X^a$ could be evaluated based on the anomalous score ($\Gamma(X^a)$) and odds ratio, $\omega(X^a,X)$.

In this work,  varieties of $\mathcal{R}(\cdot)$ are employed to select features that can be fed into $S(\cdot)$. These include \textit{Filter and Wrapper} ($R^w(\cdot)$) and\textit{ Embedded} ($R^e(\cdot)$) techniques as described below.

\subsection{Filters and Wrapper Techniques}
The first approach in our feature selection pipeline involves using filters to pre-select features and and then applying stepwise feature elimination via ordinary least squares regression as illustrated in Fig.~\ref{fig:pipeline}.  A filter-based feature selection technique ($R^w(\cdot)$) takes the set of input features $\mathcal{F}$ and provides the required number of features  $\mathcal{F}^r$ by first applying statistical methods to encode pairwise feature relations using a function $\zeta(\cdot)$, and then quantifying feature-outcome relations using a function $\tau(\cdot)$.  Varieties of $\zeta(\cdot)$ are applied based on the  type of features. To this end, features in $\mathcal{F}$  are categorised as either \textit{continuous} ($\mathcal{F}^c$ ) or \textit{categorical} ($\mathcal{F}^g$).
%The statistical methods applied to encode the significance of pairwise feature relationships are the Pearson correlation coefficient\cite{benesty2009pearson} ($\rho$) for $\mathcal{F}^c$, and  Chi-Square test~\cite{mchugh2013chi} ($\chi$)  for $\mathcal{F}^g$. Subsequently, in order to evaluate the strength of the association when statistical significance has been obtained, we employed Variance Inflation Factor~\cite{} ($\Upsilon$) for  $\mathcal{F}^c$  and Cramer's V~\cite{akoglu2018user} ($\Phi$) for $\mathcal{F}^g$, i.e., $\zeta^c(\mathcal{F}^c_{ij})= \Upsilon(\Psi(\mathcal{F}^c_i, \mathcal{F}^c_j))$. Similarly, $\zeta^g(\mathcal{F}^g_{ij})= \Phi(\chi(\mathcal{F}^g_i, \mathcal{F}^g_j))$. 
For $\mathcal{F}^c$, we use the Pearson correlation coefficient\cite{benesty2009pearson}, $\rho$, to measure the strength of a linear association between a pair of continuous features, i.e., $\rho(\mathcal{F}^c_i, \mathcal{F}^c_j)$. If $\rho(\mathcal{F}^c_i, \mathcal{F}^c_j)$ is above a given threshold, we compute the correlation between each feature in the pair and the outcome and select the one with the higher correlation. We also use the Variance Inflation Factor \cite{o2007caution} $\Upsilon$, to measure the degree of multi-collinearity between a continuous feature and other continuous features, i.e., $\Upsilon(\mathcal{F}^c_i$, $\mathcal{F}^c_{-i})$. 
For $\mathcal{F}^g$, we use the Chi-Square test~\cite{mchugh2013chi} to measure the significance of the association between two features, i.e., $\chi(\mathcal{F}^g_i, \mathcal{F}^g_j)$. Subsequently, we use the Cramer's V~\cite{akoglu2018user} to evaluate the strength of association between the two variables, i.e, $\Phi(\mathcal{F}^g_i, \mathcal{F}^g_j)$, and if above a given threshold, we use Mutual Information, $\Lambda$, to select the feature that has more mutual dependence with the target variable. 
%Consequently, $\zeta^c(\mathcal{F}^c_{ij})= \Upsilon(\rho(\mathcal{F}^c_i, \mathcal{F}^c_j))$, and $\zeta^g(\mathcal{F}^g_{ij})= \Phi(\chi(\mathcal{F}^g_i, \mathcal{F}^g_j))$.
%$\zeta_t^c$ and $\zeta_t^g$ represent the sets of continuous and categorical features that satisfy the thresholds related to the statistical significance and strengths of the relationships between features. 
Consequently, we obtain $\zeta_t^c$ and $\zeta_t^g$, the sets of continuous and categorical features that satisfy the thresholds related to the statistical significance and strengths of the relationships between features. 

%Next, wrapper method, $\tau(\cdot)$, is applied to evaluate the relationship between the filtered features ($\zeta_t^c$ and $\zeta_t^g$ ) and the outcome label $y_i$.  $\tau^c=\Psi(\zeta_t^c)$ is applied for filtered continuous ($\zeta_t^c)$) and $\tau^g=\Lambda(\zeta_t^c)$ is applied for filtered categorical $\zeta_t^g$ features, where $\Lambda$ represents the \textit{Mutual Information} value. This results  $\tau^g_t$ and $\tau^c_t$ feature sets that satisfy the thresholds. Finally, we apply a wrapper method, $\tau(\cdot)$ that uses backward elimination feature selection to obtain the required $K$ features from $\tau^g_t \cup \tau^c_t$.  Specifically, the wrapper technique involves fitting a linear model (e.g., Ordinary Least Squares) and drops less significant features recursively until a stopping criterion (until $K$ features remaining from $\tau^g_t \cup \tau^c_t$, i.e., $\mathcal{F}^r$) is met.  

Finally, we apply a wrapper method, $\tau(\cdot)$ that uses backward elimination feature selection to obtain the required $K$ features from $\zeta_t^c$ and $\zeta_t^g$.  Specifically, the wrapper technique involves fitting a linear model (e.g., Ordinary Least Squares) and drops less significant features recursively until a stopping criterion (until $K$ features remaining from $\zeta_t^c \cup \zeta_t^g$, i.e., $\mathcal{F}^r$) is met.

\subsection{Embedded Techniques}
The second approach of our feature selection pipeline (see Fig.~\ref{fig:pipeline}) is embedded based technique ($R^e(\cdot)$). This approach employs varieties of tree-based classifiers, i.e., Catboost~\cite{dorogush2018catboost} ($R^e_c$) and XGBoost~\cite{chen2016xgboost} ($R^e_x$), which  have the ability to identify duplicate features in the set and drop them without the need of using filters. A user might pre-define to use one of these classifiers or a combination of both ($R^e_{cx}$). After training of these classifiers using $\mathcal{D}  = \{(x_i,y_i)\}$, importance of features are extracted  per each  classifier, resulting $\Theta^c = [\theta_1^c,\theta_2^c,\cdots, \theta_M^c]$ and $\Theta^x = [\theta_1^x,\theta_2^x,\cdots, \theta_M^x]$ from  $R^e_c$ and $R^e_x$, respectively.  Accordingly, the top $K$ features could be extracted by ranking $\Theta^c$ and $\Theta^x$ in descending order.  In cases where a combination of the two classifiers ($R^e_{cx}$) is employed, a committee vote is applied to merge  $\Theta^c$ and $\Theta^x$  into unified $\Theta^{cx}$ ranking. To this end, MinMax~\cite{} normalization of model-specific ranking is done, resulting $\bar{\Theta}^c$  and $\bar{\Theta}^x$, e.g., $\bar{\Theta}^c = \frac{\Theta^c}{Max(\Theta^c) - Min(\Theta^c)}$. After normalisation,  $\Theta^{cx}$ is obtained by averaging $\bar{\Theta}^c$  and $\bar{\Theta}^x$ and top $K$ features ($\mathcal{F}^r$) are extracted by ranking $\Theta^{cx}$ in descending order.

% Three Embedded Techniques are implemented in the pipeline: fitting a Catboost model, fitting an Xgboost Model and fitting both Catboost and Xgboost models. These models are tree-based and have the ability to identify duplicate features in the set and drop them without the need of using filters. [citation] 

% The pipeline is designed to allow a user decide whether to fit a single model or a combination of both models to select the top K optimal features. When fitting using a single model, the feature importance of the fed covariates are calculated by the specified model and returned as a ranking list by the pipeline. The list is used to select the top K requested features based on the rankings. 

% When using both models to select features, feature importance of both models is calculated whilst fitting the models and returned as two ranked lists. A committee vote is implemented to create a new ranking list. The committee vote works as stated below:

% \begin{enumerate}
%     \item If a feature f appears in only one ranking list at position i, penalize the feature by a decided value P. The new position value is i + P.
%     \item If a feature f appears in both ranking lists at position i and j, add the positions from both lists and create a new position value for f as i + j.
%     \item Create a new list from the created positions in ascending order.
% \end{enumerate}

% The requested top k features in a dataset are sliced from the committee vote ranking list and returned as a list. 

\subsection{Multi-Dimensional Subset Scanning (MDSS)}

We employ the Multi-Dimensional Subset Scanning(MDSS)~\cite{neill2012fast,mcfowland2018efficient} from the anomalous pattern detection literature in order to validate the above automated feature selection techniques to identify differentiated patterns of care.
Subset scanning could be posed as a search problem over possible subsets in a multi-dimensional array, and it aims to identify differentiated (anomalous) subsets $X^a \subseteq \mathcal{D}$ by searching across all possible combination of covariates obtained from the previous selection stages.  Automatic stratification (AutoStrat) is a version of MDSS where the deviation between average outcomes in $\mathcal{D}$ ($\alpha_D$) and each sample ($\alpha_i$) is evaluated by maximizing a Bernoulli likelihood ratio scoring statistic, $\Gamma(S)$. The null hypothesis assumes that the likelihood of the outcome in each sample $x_i^r \in \mathcal{D}^r$  similar to expected ($\alpha_g$), i.e., $H_0: odds(y_i)=\frac{\alpha_g}{1-\alpha_g}$; while the alternative hypothesis assumes a constant multiplicative increase in the outcome odds for some given subgroup, $H_1: odds(y_i)=q\frac{\alpha_g}{1-\alpha_g}$ where $q>1$. The scoring function, $\Gamma(S)$ for a subset, $S$, which contains $N_S$ samples, is computed as:
\[
\Gamma(S) = \max_q log(q)\sum_{i\in S} y_i - N_S * log(1-\alpha_g + q\alpha_g)
\]
Consequently, subsets in which average of $y_i$ ($\alpha_y$) is greater than $\alpha_g$ will have higher scores.  Subset selection is iterated until convergence to a local maximum is found, and the global maximum is subsequently optimized using multiple random restarts.  Once the differentiated subset of samples $X^a$ is identified using MDSS, empirical p-value (via randomisation testing) is computed to evaluate the significance of the differentiation. Moreover, subset characterisation is conducted to provide interpretation of these anomalous features $\mathcal{F}^a$ and subset $X^a$.

\section{Experiments and Results}
% - Descriptive statistics (mimic) ? 
% - Feature selection outcomes and performance metrics (AUC, and other metrics)
% - DPoC results (scores, p-values, subsets, overlaps, plots for different k)
% - How to put all figures in a single image as sub images
\subsection{Dataset Used, Cohort, Covariates, and Outcomes}
The Dataset used in this project was the MIMIC-III (‘Medical Information Mart for Intensive Care’) dataset\cite{johnson_2016_data}. MIMIC-III is a freely accessible critical care dataset recording vital signs, medications, laboratory measurements, observations and notes charted by care providers, fluid balance, procedure codes, diagnostic codes, imaging reports, hospital length of stay, survival data and other data collected from over 46,000 intensive Care Unit (ICU) patients. We selected a study cohort of adult patients (16 years or older) who were admitted to the ICU for the first time, where  the length of stay was greater than 1 day, and with no hospital readmissions, no surgical cases, and having at least one chart events. The final cohort consisted of 18,761 patients. We constructed 42 covariates based on observations made on the first 24 hours of ICU admission. These included 15 numerical features, 22 binary features, and 5 nominal features. We defined the target outcome as a binary indicator variable $y_i$ such that  $y_i=1$ for patients who who died within 28 days of the onset of their ICU admission, and $y_i=0$ otherwise. 

\subsection{Feature Selection Process}
% Using the Feature Selection Pipeline, all the four techniques ( fitting a catboost Model, fitting an xgboost Model, fitting both xgboost and catboost Model and using the filters and wrapper technique) were run and four ranked lists of top K = 10 selected features extracted.
%An interesting question to know the optimal K value of features to select in a dataset was experimented by selecting top K=30 Features in the MIMIC-III dataset. These features were then run in increments of 5 in the DPoC codebase and evaluated to get the optimal K value in the MIMIC-III dataset.

%Using the MIMIC-III dataset, four sets of top K = 10 features were selected using the previously explained feature selection techniques in the pipeline. The selected features for the techniques had overlaps which can be quantified in the committee vote. A penalty P of 20 was used when calculating the committee vote. Seven features were similar from the Top K=10 selected features by xgboost and catboost separately. 
%The F1 scores of the fitted catboost model was 0.910 and the xgboost model 0.909, which are relatively high indicating that the classifier did a good job classifying patients who had either passed on or not after 28-days of admission in critical care units.

We used the 4 different techniques in our feature selection pipeline: a Filter+Wrapper method, an XGBoost-based Embedded method, a Catboost-based Embedded method, and committee vote approach that combined the rankings from the XGBoost and Catboost rankings. For each method, the our first goal was to identify the top $K=10$ features in our analytic dataset. Our second goal was to investigate the optimal K value for use in differentiated pattern detection. To do so, we used the methods to identify the top K features where K $\in$ \{5, 10, 15, 20, 25, 30\}. Table \ref{selected-features} lists the top $K=10$ features selected by the different techniques in our pipeline. The F1 scores of the fitted XGBboost and Catboost models were 0.909 and 0.910 respectively. This indicates that both classifiers performed well. Seven features were similar from the Top K=10 selected features by XGBoost and Catboost separately. 

\begin{table}[ht]
  \caption{Top 10 feature selected by the feature selection pipeline }
  \label{selected-features}
  \centering
  \begin{tabular}{p{0.1\linewidth} p{0.85\linewidth}}
    \toprule
    \textbf{Wrapper} & fluid electrolyte, liver disease, cardiac arrhythmias, coagulopathy, vent first hour, deficiency anemias, congestive heart failure, alcohol abuse, angus, hypertension \\
    \midrule
    \textbf{XGBoost} & SpO2 Mean, TempC Mean, DiasBP Mean, RespRate Mean, EndoTrachFlag, diabetes complicated, SysBP Mean, chronic pulmonary, day  name icu intime, PLATELET first  \\
    \midrule
    \textbf{Catboost} & day name icu intime, EndoTrachFlag, marital status, angus, ethnicity, SpO2 Mean, DiasBP Mean, liver disease, vent first hour, peripheral vascular disease \\
    \midrule
    \textbf{Committee Vote} & SpO2 Mean, EndoTrachFlag, DiasBP Mean, day name icu intime, TempC Mean, marital status, RespRate Mean, angus, ethnicity, diabetes complicated \\
    \bottomrule
  \end{tabular}
\end{table}

\subsection{Differentiated Pattern Detection}
%The four previously selected ranked top K lists from different feature selection techniques, together with all the 42 MIMIC-III covariates were then passed in the existing DPoC codebase separately. After the numerical features were discretized, Effect measures (Odds Ratio, Confidence interval, p\_values), most anomalous subsets, most anomalous sub-populations and scores were recorded for each of the set of features. 
We used the multidimensional subset scanning (MDSS) approach to identify the highest-scoring (most anomalous) subset when given the full set of 42 features in our analytic dataset and when given each of the top K $\in$ \{5, 10, 15, 20, 25, 30\} features selected by the four feature selection techniques in our pipeline. A total of 25 scans were conducted and compared. Before each scan, all numerical features were discretized. For each scan, we defined the observed outcome as a binary indicator of mortality within 28 days of ICU admission, and the expected outcome as a simple mean of the observed outcome. Consequently, each scan maximized the MDSS scoring function over the feature values of the input features, to identify the highest-scoring subset with the strongest evidence of the likelihood of mortality within 28 days in the subset being higher than what is expected in the sampled population. After each scan, we estimated the statistical significance of the highest-scoring subset by computing its empirical p-value via parametric bootstrapping. 

\begin{figure}[ht]
  \centering
  \fbox{\includegraphics[scale=0.3]{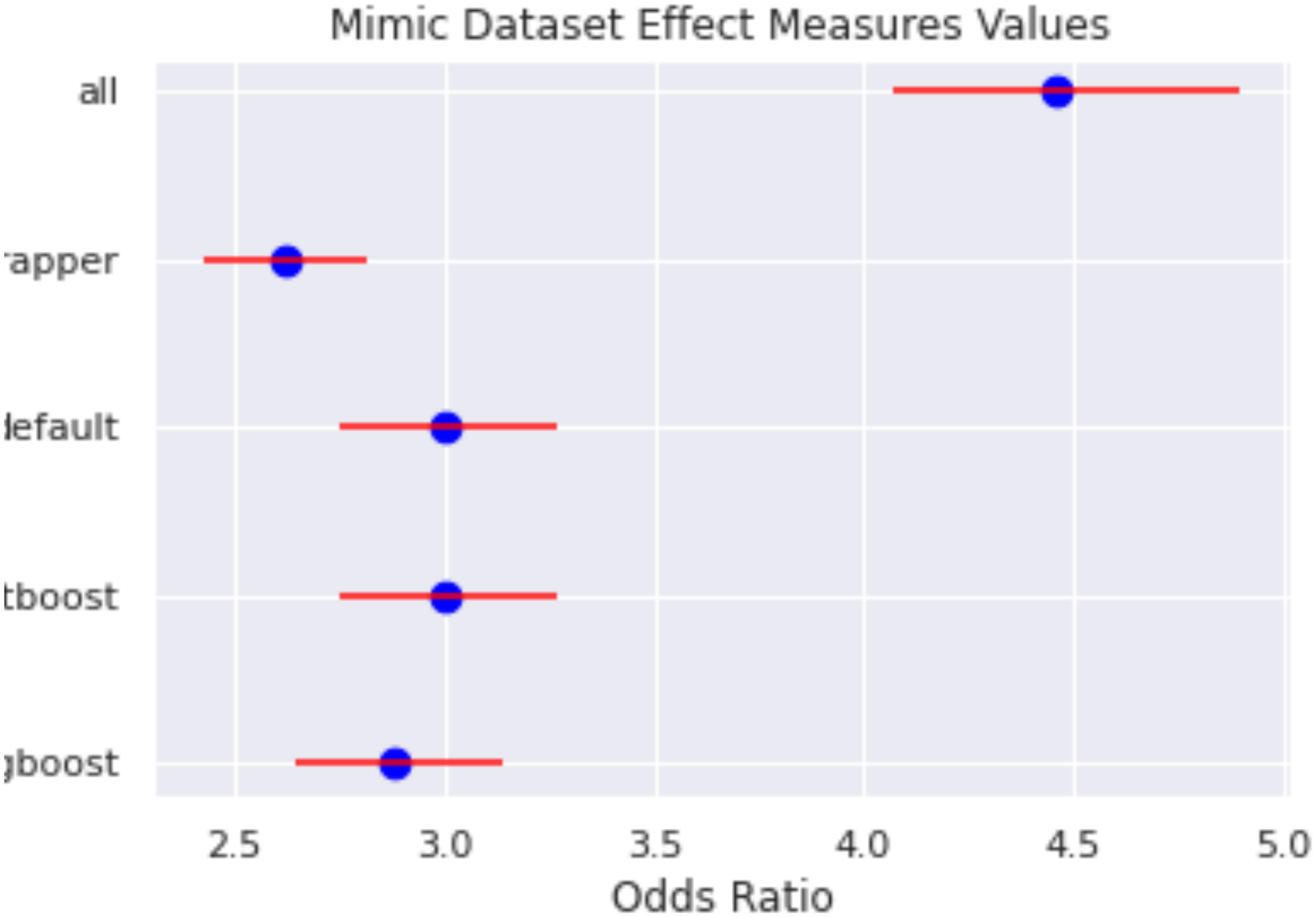}}
  \caption{Odds Ratio comparison among feature Selection techniques and whole MIMIC-III dataset}
  \label{fig:odds_ratio_comparison}
\end{figure}

The results of each scan included the highest-scoring subset, the score of the highest-scoring subset, the empirical p-value of the score, the subpopulation of patients in the highest-scoring subset, and the ratio of the odds of the outcome in the anomalous subpopulation to the odds of the outcome in the complement subpopulation. 
We found that scanning over all the features in the analytic dataset generated the highest score and the and odds ratio. Figure \ref{fig:odds_ratio_comparison} illustrates this by comparing the odds ratio statistics after subset scanning over all features compared to scanning over top $K=10$ features selected by different techniques. 

%These observations suggest that using the whole dataset to perform the multi-dimensional scanning would lead to better anomalous subsets. However, with the objective of the project being able to reduce the time and resources the multi-dimensional scan runs for and consumes; it is important to find an optimal value of K selected features where the scores and the Odds Ratio are equal or almost equal to all the covariates in the dataset.
Of note is that the $95\%$ confidence interval of odds ratio when scanning over all features is higher and does not overlap with any of the $95\%$ confidence intervals of the odds ratios when scanning over the top $K=10$ features selected by the different techniques. Interestingly, the $95\%$ confidence intervals of the odds ratios of the different techniques have considerable degrees of overlap implying that they are not truly different from each other. Assuming that scanning over all features discovers a truly anomalous subpopulation, these findings suggest that the supervised feature selection processes used in this study can introduce biases in anomalous pattern detection that are characterized by different compositions of the anomalous subsets and lower measures of effect. Consequently, it is important to find the optimal value of K that minimizes this bias such that scores and measures of effect obtained when using a feature selection technique approach those obtained when scanning over all features.

\begin{figure}[ht]
  \centering
  \fbox{\includegraphics[scale=0.3]{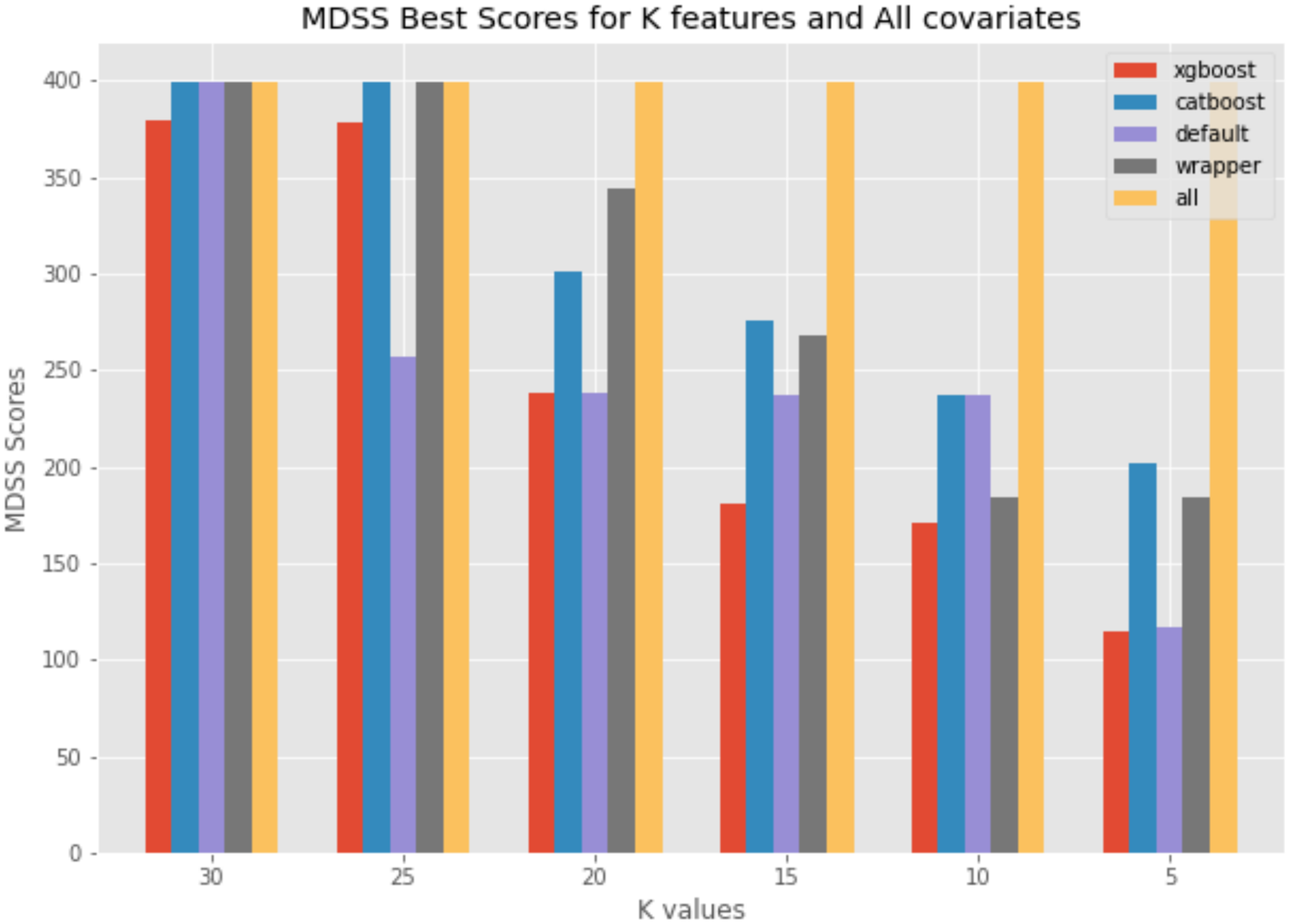}}
  \caption{Optimal Value of K in the MIMIC-III dataset}
  \label{fig:optimal_k_scores}
\end{figure}

After running the MDSS experiment to get the optimal value of K selected features, the results in figure \ref{fig:optimal_k_scores} suggested that it was possible to attain the same scores as all the covariates in a dataset using less features with K = 25 saving on resources and time to run the MDSS scan. The feature selection technique is heuristic and is dependent on the dataset for the optimal value of K. 

%Figure \ref{fig:optimal_k_scores} illustrates the the score of the highest-scoring subset after scanning over all features in the analytic top K features selected by different targets/

% P-values of the selected K features ..... 
% Code did not run. Error
%Other publicly available datasets such as the 30-days Diabetes Readmission dataset and the Star Tennessee dataset were used in the same experiment to validate results and also to illustrate that the feature selection pipeline can handle any dataset. 

\section{Conclusions and Future Work}
% -summary of the study and its findings
% - the key result (feature selection)
% - another result (dpoc)
% - limitations of the study 

This work proposes a supervised automated feature selection pipeline to select optimal K features for the multi-dimensional subset scanning project. From this work, it is observed that finding an optimal value of K selected features for any dataset saves on resources rather than scanning through the whole dataset containing possible redundant covariates in the feature space.

The proposed supervised automated feature selection pipeline exhibits an optimal way for a user to select features by using one or more feature selection techniques. The various feature selection techniques can be considered optimal as the selected features overlap across the techniques.

Using the proposed supervised automated feature selection techniques succoured us achieve the main objective of using less K features in the Multi-Dimensional Subset Scanning pipeline and achieving equally better scores as using all N features of the dataset.  

% Our key contributions in this paper is the proposed supervised feature selection pipeline that allows a dataset with N features to have optimal K selected features that perform equally as using the whole N features of the dataset. 

The limitation of the study is that the multi-dimensional subset scanning is an unsupervised machine learning technique and the feature selection pipeline is based on a supervised technique. Setting of several thresholds in the filtering technique also introduces a limitation in our study given that these thresholds have to be dependent on the datasets used.

% Future work could involve testing out unsupervised feature selection techniques to validate the results.

We propose future work to explore how unsupervised feature selection techniques perform in relation to our supervised approach. Also, the use of thresholds in our work introduces an extra step to get the appropriate threshold for each dataset. This process could also be automated. We also intend to evaluate the runtime performances of the supervised feature selection techniques and compare these with unsupervised feature selection approaches.

\bibliographystyle{unsrt}
\bibliography{neurips_2021}

\begin{thebibliography}{10}

\bibitem{appleby2011variations}
John Appleby, Veena Raleigh, Francesca Frosini, Gwyn Bevan, Haiyan Gao, and Tom
  Lyscom.
\newblock Variations in health care.
\newblock {\em The good, the bad and the inexplicable. London: The King's
  Fund}, 2011.

\bibitem{krumholz2013variations}
Harlan~M Krumholz.
\newblock Variations in health care, patient preferences, and high-quality
  decision making.
\newblock {\em Journal of the American Medical Association}, 310(2):151--152,
  2013.

\bibitem{senn2016mastering}
Stephen Senn.
\newblock Mastering variation: variance components and personalised medicine.
\newblock {\em Statistics in Medicine}, 35(7):966--977, 2016.

\bibitem{mcfowland2018efficient}
Edward McFowland~III, Sriram Somanchi, and Daniel~B Neill.
\newblock Efficient discovery of heterogeneous treatment effects in randomized
  experiments via anomalous pattern detection.
\newblock {\em arXiv preprint arXiv:1803.09159}, 2018.

\bibitem{neill2012fast}
Daniel~B Neill.
\newblock Fast subset scan for spatial pattern detection.
\newblock {\em Journal of the Royal Statistical Society: Series B (Statistical
  Methodology)}, 74(2):337--360, 2012.

\bibitem{zhang2016identifying}
Zhe Zhang and Daniel~B Neill.
\newblock Identifying significant predictive bias in classifiers.
\newblock {\em arXiv preprint arXiv:1611.08292}, 2016.

\bibitem{somanchi2017}
Edward Somanchi, Sriram McFowland~III and Daniel~B Neill.
\newblock Detecting anomalous patterns of care using health insurance claims.
\newblock {\em Presented at Conference on Information Systems and Technology},
  2017.

\bibitem{ogallo2021detection}
William Ogallo, Girmaw~Abebe Tadesse, Skyler Speakman, and Aisha
  Walcott-Bryant.
\newblock Detection of anomalous patterns associated with the impact of
  medications on 30-day hospital readmission rates in diabetes care.
\newblock In {\em AMIA Annual Symposium Proceedings}, volume 2021, page 495,
  2021.

\bibitem{mcfowland2013fast}
Edward McFowland, Skyler Speakman, and Daniel~B Neill.
\newblock Fast generalized subset scan for anomalous pattern detection.
\newblock {\em The Journal of Machine Learning Research}, 14(1):1533--1561,
  2013.

\bibitem{miao_2016_a}
Jianyu Miao and Lingfeng Niu.
\newblock A survey on feature selection.
\newblock {\em Procedia Computer Science}, 91:919--926, 2016.

\bibitem{kumar_2015_a}
Amit Kumar.
\newblock A survey on feature selection algorithms.
\newblock {\em International Journal on Recent and Innovation Trends in
  Computing and Communication}, 3:1895--1899, 2015.

\bibitem{molina_2002_feature}
L.C. Molina, L.~Belanche, and A.~Nebot.
\newblock Feature selection algorithms: a survey and experimental evaluation.
\newblock {\em Proceedings of the IEEE International Conference on Data
  Mining}, 2002.

\bibitem{benesty2009pearson}
Jacob Benesty, Jingdong Chen, Yiteng Huang, and Israel Cohen.
\newblock {\em Noise reduction in speech processing}, volume~2.
\newblock Springer Science \& Business Media, 2009.

\bibitem{o2007caution}
Robert~M O’brien.
\newblock A caution regarding rules of thumb for variance inflation factors.
\newblock {\em Quality \& Quantity}, 41(5):673--690, 2007.

\bibitem{mchugh2013chi}
Mary~L McHugh.
\newblock The chi-square test of independence.
\newblock {\em Biochemia Medica}, 23(2):143--149, 2013.

\bibitem{estevez2009normalized}
Pablo~A Est{\'e}vez, Michel Tesmer, Claudio~A Perez, and Jacek~M Zurada.
\newblock Normalized mutual information feature selection.
\newblock {\em IEEE Transactions on Neural Networks}, 20(2):189--201, 2009.

\bibitem{akoglu2018user}
Haldun Akoglu.
\newblock User's guide to correlation coefficients.
\newblock {\em Turkish Journal of Emergency Medicine}, 18(3):91--93, 2018.

\bibitem{dorogush2018catboost}
Anna~Veronika Dorogush, Vasily Ershov, and Andrey Gulin.
\newblock Catboost: gradient boosting with categorical features support.
\newblock {\em arXiv preprint arXiv:1810.11363}, 2018.

\bibitem{chen2016xgboost}
Tianqi Chen and Carlos Guestrin.
\newblock Xgboost: A scalable tree boosting system.
\newblock In {\em Proceedings of the ACM SIGKDD International Conference on
  Knowledge Discovery and Data Mining}, pages 785--794, 2016.

\bibitem{johnson_2016_data}
Alistair Johnson, Tom Pollard, Lu~Shen, Li-Wei Lehman, Mengling Feng, Mohammad
  Ghassemi, Benjamin Moody, Peter Szolovits, Leo Celi, and Roger Mark.
\newblock Data descriptor: Mimic-iii, a freely accessible critical care
  database.
\newblock {\em Scientific Data}, 3, 05 2016.

\end{thebibliography}

% 	\bibliographystyle{IEEEtran}
% 	\bibliography{neurips_2021}
\end{document}